\pdfoutput=1

\documentclass[11pt]{article}

\usepackage[]{EMNLP2023}

\usepackage{times}
\usepackage{latexsym}
\usepackage{graphicx}

\usepackage[T1]{fontenc}

\usepackage[utf8]{inputenc}

\usepackage{microtype}

\usepackage{inconsolata}

\title{Mavericks at NADI 2023 Shared Task: Unravelling Regional Nuances through Dialect Identification using Transformer-based Approach}

\author{Vedant Deshpande \thanks{~ Equal contribution}~~, 
Yash Patwardhan$\footnotemark[1]$~, 
Kshitij Deshpande$\footnotemark[1]$~, \\
{\bf Sudeep Mangalvedhekar}$\footnotemark[1]$~ \and
{\bf Ravindra Murumkar}$\footnotemark[1]$~ \\
Pune Institute of Computer Technology, Pune \\
\texttt{\{vedantd41, yash23pat, kshitij.deshpande7, sudeepm117\}@gmail.com,}\\
\texttt{rbmurumkar@pict.edu}}

\begin{document}
\maketitle
\begin{abstract}
In this paper, we present our approach for the "Nuanced Arabic Dialect Identification (NADI) Shared Task 2023". We highlight our methodology for subtask 1 which deals with country-level dialect identification. Recognizing dialects plays an instrumental role in enhancing the performance of various downstream NLP tasks such as speech recognition and translation. The task uses the Twitter dataset (TWT-2023) that encompasses 18 dialects for the multi-class classification problem. Numerous transformer-based models, pre-trained on Arabic language, are employed for identifying country-level dialects. We fine-tune these state-of-the-art models on the provided dataset. The ensembling method is leveraged to yield improved performance of the system. We achieved an $F_1$-score of 76.65 (11th rank on the leaderboard) on the test dataset.

\end{abstract}

\section{Introduction}
Dialects, which are variations of a language, often differ in their vocabulary, grammar, pronunciation, and occasionally even cultural quirks. The practice of identifying the particular dialect or regional variety of a language that is used in a text or speech sample is known as dialect identification. The goal of dialect identification is to categorize a text or speech into one of the many dialects or regional adaptations that may exist. For many NLP applications, including language modeling, speech recognition, and data retrieval, this task may be vital.

Arabic, with its plethora of dialects, is a rich language. However, many of these dialects are not studied in depth because of a dearth of monetary backing and available datasets. Arabic dialect identification can assist in perpetuating linguistic diversity by acknowledging and valuing various dialects. It contributes to addressing the gap between existing NLP techniques and the rich fabric of regional dialectal differences in a globalized setting. Rule-based strategies for Arabic dialect identification have given way to data-driven techniques, with a focus on machine learning, deep learning, and the creation of corpora of languages and datasets. The accuracy of dialect detection has risen significantly with the use of multilingual pre-trained models such as BERT and its derivatives.

This paper presents our approach for subtask 1: Country-level Dialect Identification, which poses a multiclass classification problem \cite{abdul-mageed-etal-2023-nadi}. Multiclass classification is a form of statistical modeling or machine learning problem where the objective is to classify data into more than two unique classes or labels. We aim to classify the tweets and map them into their respective dialect labels. We have demonstrated the use of various transformer-based models on the given Arabic data. The ensembling method has been leveraged to enhance the performance of the proposed system.

\begin{figure*}[t]
    \centering
    \includegraphics[width=\textwidth]{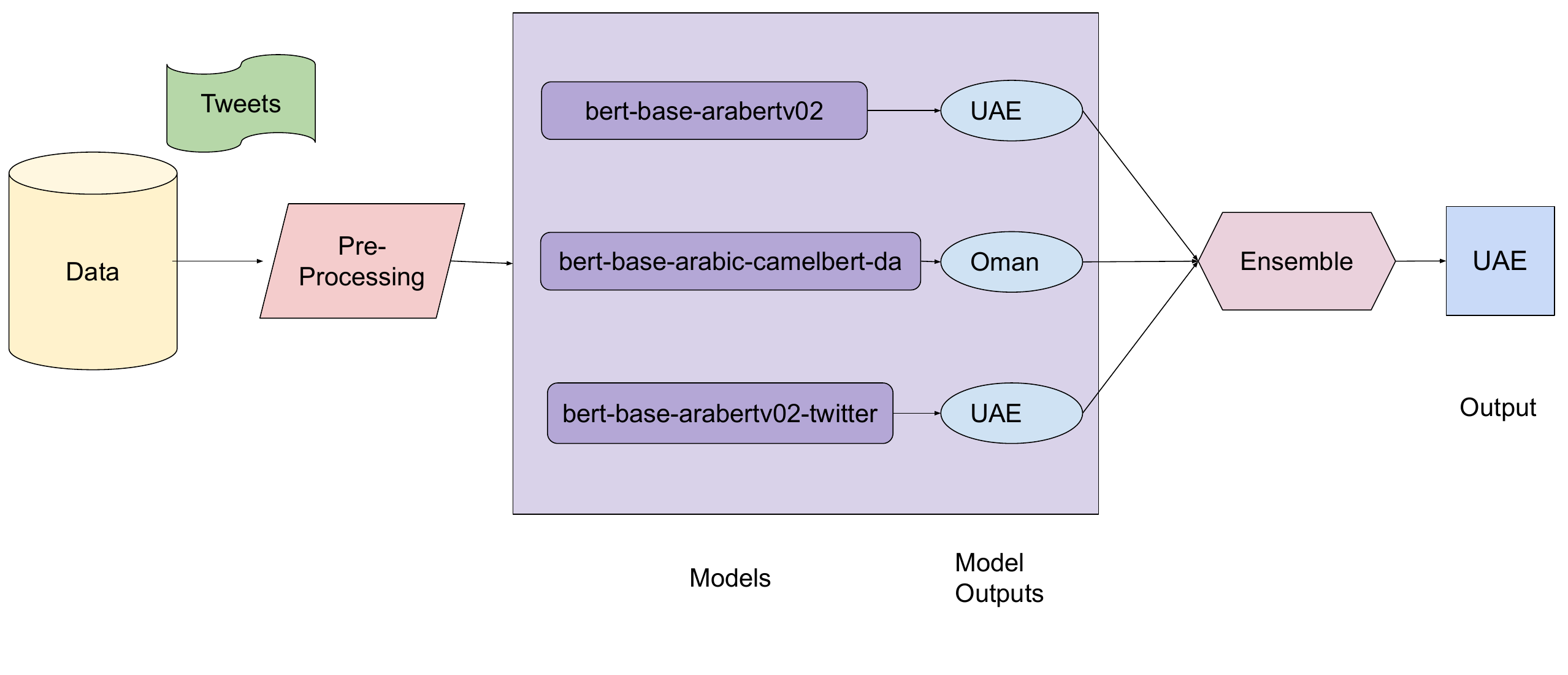}
    \caption{System architecture}
    \label{fig:ensemble}
\end{figure*}

\section{Related Work}
Dialect detection in Arabic is an arduous task due to several factors, including the lack of a consistent spelling system, the medium's characteristics, and the scarcity of data. 
Surveys on deep learning and Natural Language Processing methods for processing Arabic data were presented in 2015  \cite{shoufan2015} and 2017 \cite{ALAYYOUB2018522}, focusing on the identification of Arabic dialects.
However, only 6 Arabic dialect classes had been examined until that time. 
The MADAR project was launched in 2018 to provide a large corpus of 25 Arabic city dialects \cite{bouamor-etal-2018-madar}.
A study on the classification of dialects in 25 Arab cities used multi-label classification methods and examined a wide range of features, yielding promising results \cite{salameh2018}.
Employing supervised machine learning methods on Arabic NLP tasks was found to be a difficult feat because of the lack of resources in the Arabic language \cite{el-mekki-etal-2020-weighted}.
As a result, scholars and researchers have introduced plenty of initiatives to make new datasets available and encourage more people to work in the field of Arabic NLP. One of the initiatives, Nuanced Arabic Dialect Identification (NADI) shared tasks, was started in 2020 which comprised country-level and province-level dialect detection \cite{abdul-mageed-etal-2020-nadi}.
BERT (Bidirectional Encoder Representation from Transformers) \cite{devlin-etal-2019-bert} models have been commonly used 
for these dialect detection tasks.
A multilingual BERT model was pretrained on unlabeled tweets and fine-tuned for the classification task by \citet{mansour-etal-2020-arabic}.
AraBERT was finetuned on an additional dataset produced by reverse translating the NADI dataset and employed for the dialect detection task by \citet{tahssin-etal-2020-identifying}.
Furthermore, \citet{gaanoun-benelallam-2020-arabic} utilized ensembling methods and semi-supervised methods along with Arabic-BERT.
A system comprising an ensembling of multiple models was created using MARBERT as the base model, which yielded promising results \cite{alkhamissi-etal-2021-adapting}. 
In the NADI 2022 shared task, \citet{alshenaifi-azmi-2022-arabic} pretrained AraBERT model and BiLSTM model for dialect detection. Various models were combined and performance was enhanced using a combination of TF-IDF and n-grams. 
An ensembling of transformer-based models, predominantly using variations of MARBERT was employed for dialect detection as well as sentiment analysis, in the NADI 2022 shared tasks \cite{bayrak-issifu-2022-domain}, \cite{khered-etal-2022-building}.
\cite{oumar-mrini-2022-ahmed} addressed the issue regarding an imbalance in the classes of the NADI dataset by using focal loss and employed various Arabic BERT-based models.

This paper proposes a system that employs an ensemble of transformer-based models, specifically variations of BERT for the classification task. 

\renewcommand{\arraystretch}{1.5}
\begin{table}[ht]
\begin{center}

\begin{tabular}{|c|c|}
\hline
\textbf{Dataset} & \textbf{Number of Samples} \\ \hline
Training & 18000 \\ \hline
Development & 1800 \\ \hline
Testing & 3600 \\ \hline
\end{tabular}
\caption{Dataset's training, development, and test split}
\label{Table:1}
    
\end{center}
\end{table}

\section{Data}

The dataset provided for subtask 1: Country-level dialect identification contains tweets. The given Twitter dataset comprises 18 dialects and a corpus of a total of 23400 tweets. The entire dataset is split into training (76.92\%), development (7.69\%), and test (15.38\%). Additionally, datasets of previous years \cite{abdul-mageed-etal-2020-nadi, abdul-mageed-etal-2021-nadi, bouamor-etal-2018-madar} are also provided for the training purpose. As shown in table \ref{Table:1}, the training data has 18000 tweet samples, development data has 1800 samples and testing data has 3600 samples. The dataset contains features such as id, content, and label. Every sample's tweet content in the training dataset is labeled with its dialect. This subtask falls under the category of multi-class classification.

The provided dataset needed to be pre-processed before passing it to the model. We make use of regular expressions to remove "noisy" elements from the input texts. Texts like "USER", "NUM" and "URL" are removed from the input because they don't contribute additional information to the model's understanding.

\section{System}
The given subtask tackles the problem of country-level dialect identification. This comes under the umbrella of multi-class classification problems for which Language Models have been extensively used and have achieved impressive results. The models are trained for 10 epochs with a learning rate of 1e-5, a batch size of 32, and the AdamW optimizer. We experiment and use several language models and ensembling methods in our research, as shown in Figure \ref{fig:ensemble}.

\subsection{AraBERT}
\citet{antoun-etal-2020-arabert} addresses how BERT models that have been pre-trained on a sizable corpus of a particular language, such as Arabic, do well on language comprehension tasks. They point out several such models, which are used in our study to help deliver cutting-edge outcomes for the Arabic language.

The 70 million phrases that make up the pre-training dataset, which is around 24 GB in size, are used to train the models. The news in the data covers a wide range of topics that is valuable for many downstream applications. The pre-training tasks that aid in the models' contextual knowledge of the input sequence include the Next Sentence Prediction Task and Masked Language Modelling Tasks. To demonstrate AraBERT's efficacy across diverse tasks and domains, it was tested on three NLP tasks: entity recognition, sentiment analysis, and question-answering.

Small adjustments have been made to the pre-training phases and parameters for the selected AraBERT model versions. AraBERT v1 or v0.1 are the original models, and v2 or v0.2 are the more recent versions with improved pre-processing and vocabulary. In addition to the dataset used for the other v0.2 models, AraBERTv0.2-Twitter-base is pre-trained with 60 million multi-dialect tweets. It possesses 136 parameters. Pre-trained examples for AraBERTv2-base include 207M instances with a sequence length of 512 and 420M examples with a sequence length of 128.

\subsection{CAMeLBERT}
\citet{inoue-etal-2021-interplay} introduced the CAMelBERT model collection, which consists of more than eight pre-trained models for NLP tasks in Arabic. 
The parameters taken into consideration for the experiment were the task type, language variant, and size. 
Language models were provided in several variants, including classical Arabic (CA), dialectal Arabic (DA), and Modern Standard Arabic (MSA), with the DA variant being chosen for this study.
The models were pretrained on variations of the MADAR dataset and NADI datasets for the Dialect Identification task.
CAMelBERT was trained with the Adam optimizer and a learning rate of 1e-4. 
The pre-trained models are evaluated on five major tasks in NLP: Sentiment Analysis, Dialect Identification, POS tagging, Named Entity Recognition, and Poetry Classification.

\section{Ensembling}
Ensembling is a technique that integrates the output of multiple models to get the system's eventual outcome. For this, both statistical and non-statistical methods are employed. Ensembling is beneficial since it contributes to the production of results that are superior to those provided by the individual models.

We note that the "hard voting" ensemble strategy emerges as the most effective and precise among the many strategies used for ensembling. In hard voting, the final prediction is chosen based on the majority vote or the "mode" of all the predictions. It reduces the volatility in the outcomes and aids in strengthening the system's robustness.

\renewcommand{\arraystretch}{1.5}
\begin{table}[ht]
\begin{center}    
\begin{tabular}{|c|c|}
\hline
\textbf{Model} & \textbf{\begin{tabular}[c]{@{}c@{}}$F_1$ Score\end{tabular}} \\ \hline 
\textbf{AraBERTv02-Twitter-base} & \textbf{77.03} \\
CAMeLBERT-DA & 72.78 \\ 
AraBERTv02-base & 73.07 \\ \hline
\textbf{Ensemble - Hard Voting} & \textbf{77.62} \\ \hline
\end{tabular}
\caption{Results for Dialect Identification Task on the Development dataset }
\label{Table:2}
\end{center}
\end{table}

\renewcommand{\arraystretch}{1.5}
\begin{table}[ht]
\begin{center}
\begin{tabular}{|c|c|}
\hline
\textbf{Model} & \textbf{\begin{tabular}[c]{@{}c@{}}$F_1$ Score\end{tabular}} \\ \hline 
\textbf{AraBERTv02-Twitter-base} & \textbf{75.17} \\
CAMeLBERT-DA & 71.99 \\ 
AraBERTv02-base & 72.09 \\ \hline
\textbf{Ensemble - Hard Voting} & \textbf{76.65} \\ \hline
\end{tabular}
\caption{Results for Dialect Identification Task on the Test dataset}
\label{Table:3}
\end{center}
\end{table}

\section{Results}
This section discusses the results obtained by our system and analyses its performance. Table \ref{Table:2}  and Table \ref{Table:3} depict our scores for the individual models used and the corresponding ensembled score on the development dataset and the test dataset respectively. The $F_1$ score is used as the official metric for scoring the systems.

AraBERTv02-Twitter-base outperforms the other models with an $F_1$ score of 77.03 on the development dataset and 75.17 on the test dataset. This performance demonstrates the benefits of utilizing a model that is pre-trained on a corpus similar to the one the task demands. AraBERTv02-Twitter-base is pre-trained on 60M multi-dialect tweets besides the usual datasets used for AraBERT models, giving it an edge over other models for this particular task. We select the ensemble-based system as our final approach since it produces outcomes with minimal variation and offers more stable predictions. This is justified by the superior performance of our system in the final evaluation stage. Our final system achieved an $F_1$ score of 76.65 on the test dataset.

\section{Conclusion}
This paper compares several transformer-based models on the task of Nuanced Arabic Dialect Identification (NADI). AraBERTv02-Twitter-base is found to outperform other models for this task. It achieves an $F_1$ score of 76.65. We use hard voting-based ensembling as the final approach for our system as it generates predictions that are stable while also improving the overall performance. 
With higher computational resources at hand, the performance of the system can be improved by training it for longer and by using bigger models for the system. Models that are specifically pre-trained on data that is similar to the data used in the task at hand can help enhance understanding and in turn, give better performance. We can also experiment with other suitable ensembling methods and gauge their efficiency for our task.

\section*{Limitations}
Models used for this task are computationally heavy and require significant computing resources for inference. As a result in certain real-world applications where there are compute constraints, using the system may pose a challenge. The data used for evaluation and pre-training of the models mentioned may have been biased even though the quality of the data used is high. Thus, it may not accurately represent real-world scenarios.


\bibliography{anthology,custom}
\bibliographystyle{acl_natbib}


%



\end{document}